\documentclass{article}
\usepackage{spconf,amsmath,graphicx}

\usepackage{hyperref}
\usepackage{color}

\title{Deep Photo Cropper and Enhancer}
\name{Aaron Ott$^{\ddag}$$^{\star}$   \thanks{$\star$ First and second authors contributed equally.}\qquad Amir Mazaheri$^{\dagger}$$^{\star}$  \qquad Niels D. Lobo$^{\dagger}$ \qquad Mubarak Shah$^{\dagger}$}
  
\address{$^{\dagger}$ \small Center for Research in Computer Vision (CRCV)-University of Central Florida ~~~
      $^{\ddag}$ North Carolina State University}

\begin{document}
%
\maketitle
%
\begin{abstract}
This paper introduces a new type of image enhancement problem. Compared to traditional image enhancement methods, which  mostly deal with pixel-wise modifications of a given photo, our proposed task is to crop an image which is embedded within a photo and enhance the quality of the cropped image. We split our proposed approach into two deep networks: deep photo cropper and deep image enhancer. In the photo cropper network, we employ a spatial transformer to extract the embedded image. In the photo enhancer, we employ super-resolution to increase the number of pixels in the embedded image and reduce the effect of stretching and distortion of pixels. We use cosine distance loss between image features and ground truth for the cropper and the mean square loss for the enhancer. Furthermore, we propose a new dataset to train and test the proposed method. Finally, we analyze the proposed method with respect to qualitative and quantitative evaluations.
\end{abstract}
\section{Introduction}
\label{sec:intro}

\begin{figure*}
    \centering
    \includegraphics[width=\linewidth]{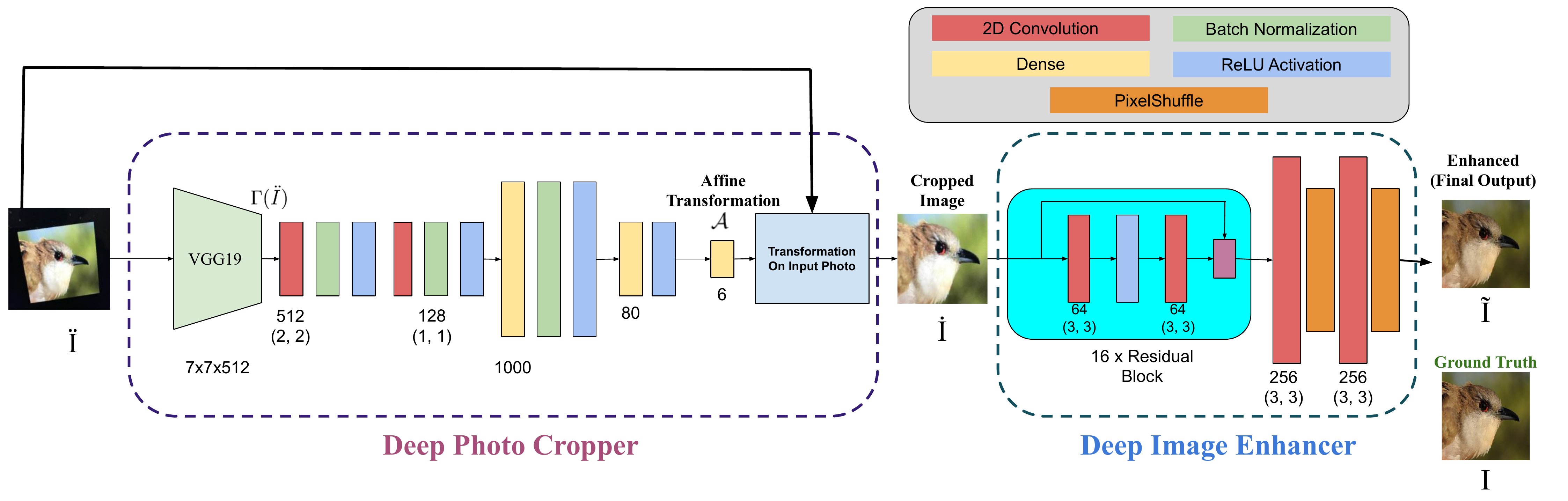}
    \caption{Given a photo of an image, which can be taken of a screen showing that image or of a wall-paper poster, our task is to crop an image that is embedded within a photo and enhance the quality of the cropped image. We propose a network divided into two sequentially connected deep sub-networks: a photo cropper and an image enhancer. We train the proposed network in an end-to-end fashion. We receive a feature map of the input image from a pre-trained VGG19~\cite{simonyan2014very} network and apply several layers until we get a 6-D  output, which represents an affine transformation $\mathcal{A}$. We then feed the affine transformation and the input photo into our Spatial Transformer to get a cropped image. We employ the Spatial Transformer Network (STN) as discussed in \cite{jaderberg2015spatial}. For the image enhancement network, we adopt an EDSR~\cite{lim2017enhanced} architecture (Enhanced Deep Super Resolution Network), which is a single-image super-resolution network. EDSR utilizes PixelShuffle~\cite{shi2016real} to increase the size of the input image, and residual blocks to enable the network to produce detailed patterns from only a few input pixels. We initialize the shown layers' weights with the pre-trained network from \cite{lim2017enhanced} and update the weights of the enhancer end-to-end with the cropper layers using our proposed loss function in Equation~\ref{eq:eloss}.}
    \label{fig:abstract}
\end{figure*}

In this paper, we address the task of Deep Image Cropping and Enhancement (DCE). DCE is related to two traditional problems in image processing: image recovery, and enhancement. In DCE, given a photo that contains an embedded image, like a photo taken from a computer  monitor that is showing e.g. an image of a bird, the goal is to recover the original bird image (see Figure~\ref{fig:abstract}).
There are many real-world use-cases for this task, particularly in situations when there is no easy access to the original version of an image such as an image printed on a wall poster, identification badge, credit card, a printed document, etc. Our proposed approach benefits from deep learning and eases the process of making a high quality digital copy of images that are printed, or  are being shown on a computer screen.  Though software that can crop and fix discoloration in images already exists, it requires extensive manual effort, and is not practical when dealing with more than a handful of images at a time. Most commercial applications use traditional approaches like Hough transformation to find the boundary of the embedded image, and color histogram normalization  to make it visually appealing (see Section~\ref{seq:Experiments} for a qualitative comparison). In this research, we employ deep learning-based approach and  develop an end-to-end image cropper and enhancer  network (see Figure~\ref{fig:abstract}).




\noindent \textbf{Contributions:}
\textbf{(a)} We tackle an advanced type of image enhancement and recovery problem which has many real world applications. To the best of our knowledge, we are the first to systematically study this problem, and present   a comprehensive solution. \textbf{(b)} We propose a novel approach to solve the introduced problem defined as cropping and enhancement sub-tasks. Our proposed deep learning based approach is end-to-end trainable, and does not need extra human interaction to perform the task. \textbf{(c)} We collect a new dataset to train and test the proposed approach, and conduct comprehensive experiments. To the best of our knowledge, this is the first available dataset to be used for our proposed problem.

\section{Related Work}
\label{seq:rel}
\textbf{Image/Video Enhancement} is one of the important problems in  Image Processing.   Photo enhancement covers many aspects such as image colorization~\cite{yoo2019coloring,messaoud2018structural,zhang2017real}, in-painting~\cite{yeh2017semantic,yang2017high,demir2018patch}, denoising~\cite{elad2006image,simonyan2014very}, reflection removal~\cite{Yang_2019_CVPR,Punnappurath_2019_CVPR}, super-resolution~\cite{lim2017enhanced,Hu_2019_CVPR}, etc. Some image enhancement tasks like image colorization and in-painting have used strong self-supervision tools for unsupervised learning \cite{pathak2016context,singh2018self}. Automatic deep photo cropping and enhancement can leverage all kinds of image enhancement techniques; however, in this research we focus on a super resolution based image enhancement architecture~\cite{lim2017enhanced} (See Section~\ref{seq:Enhancer}).


\noindent \textbf{Spatial Transformer:}
Parametric image transformations, such as affine, homography, etc., are the most fundamental tools in computer vision \cite{friston1995spatial,ashburner1997spatial}. These transformation have been successfully used in image registration~\cite{davatzikos1997spatial}, image mosaicing~\cite{tian2003comprehensive}, etc. In this paper, we are interested in using image transformation as a prepossessing step prior to cropping a photo. 
In particular, we use a spatial transformer employing an affine transformation in a deep neural network, while being able to back-propagate the gradients through the transformation matrix. The proposed formulation in~\cite{jaderberg2015spatial} allows us to have a trainable neural network that can produce a transformation matrix. Although the spatial transformer  in~\cite{jaderberg2015spatial} has been used for multiple tasks such as deformable CNNs~\cite{dai2017deformable} and Object Detection~\cite{He_2017_ICCV}, we use it in an image recovery and enhancement problem for the first time.

\section{Approach}
\label{sec:approach}
We decompose the ``Deep Cropper and Enhancement'' (DCE) task into a 
Cropper, $\mathcal{C}$, and Enhancer, $\mathcal{E}$, deep networks with parameters $\theta_\mathcal{C}$ and  $\theta_\mathcal{E}$ which are defined as follows:
\begin{equation}\label{eq:c}
\dot{I}, \mathcal{A} = \mathcal{C}(\ddot{I}, \theta_{\mathcal{C}}),
\end{equation}
and,
\begin{equation}\label{eq:e}
\Tilde{I} = \mathcal{E}(\dot{I}, \theta_\mathcal{E}).
\end{equation}
$\ddot{I} \in [0, 255]^{H\times W\times 3}$ is the RGB input photo taken by the user. $\dot{I} \in \mathcal{R}^{H \times W \times 3}$ and transformation matrix $\mathcal{A}$ are the outputs of  $\mathcal{C}$. 
The output $\ddot{I}$ of Cropper $\mathcal{C}$ is input to the Enhancer $\mathcal{E}$, which outputs the enhanced image $\Tilde{I}$. 
We learn all the parameters $\theta = [\theta_\mathcal{C}, \theta_\mathcal{E}]$ in an end-to-end fashion. 
\subsection{Deep Photo Cropper} \label{seq:DeepCropper}
The cropper, $\mathcal{C}$, 
predicts an Affine transformation matrix, $A \in \mathcal{R}^6$, which transforms the input photo.  An affine transformation can  rotate, shift, and scale the input photo to produce the cropped image. After the transformation, we crop a $224 \times 224$ block from the center of the transformed photo.

We start the cropper network by feeding the input image to a VGG19 network~\cite{simonyan2014very} pre-trained on imagenet, and extracting the features from the last pooling layer. We denote the spatial feature extracted from VGG19 by $\Gamma(\ddot{I}) \in \mathcal{R}^{7 \times 7 \times 512}$. 
We pass the $\Gamma(\ddot{I})$ into two convolution layers with 512 and 128 filters, $2\times2$ and $1\times1$ kernel sizes, and stride one. After the convolution layers, we flatten the spatial features into a universal vector $\in \mathcal{R}^{6272}$ ($7 \times 7 \times 128 = 6272$). Using three Fully-Connected (FC) layers, we first map the universal vector into $\mathcal{R}^{1000}$, and then into an $\mathcal{R}^{80}$ vector. Finally, the last FC layer produces a $\mathcal{R}^6$ vector which represents the affine transformation $\mathcal{A}$. Note that we use bias weights in all the layers, and initialize all of them except the last FC using \cite{glorot2010understanding}. However, we initialize the last FC layer weights with all zeros, and the 6 dimensional bias by flattening $\bigl[ \begin{smallmatrix}1 & 0 & 0\\ 0 & 1 & 0\end{smallmatrix} \  \bigr ]$. This way, the last layer's initial output will be an identity transformation.

Finally, we apply the produced affine transformation $\mathcal{A}$ to the input photo $\ddot{I}$ using the Spatial Transformer Network (STN) formulation provided in \cite{jaderberg2015spatial}, and crop the center $224 \times 224$ box of the photo to obtain $\dot{I} \in \mathcal{R}^{H \times W \times C}$.

 In addition to affine transformation, we examined other possible spatial transformations, such as projective transformation or homography. However, we observe that more complicated transformations make the training process harder, and the network produces poor results. Also, the embedded target image may have varying sizes in the photo. A photo that is taken from a longer/shorter distance results in a smaller/larger portion covered by the embedded image. It is very important for an automatic deep image cropper to be flexible for any range of distance. To help our model overcome these challenges, we examine applying multiple levels of spatial transformations on the input photo by stacking multiple instances of the proposed cropper module. The croppers are connected sequentially, and each instance of the cropper has a separate set of parameters. The output of the first cropper is connected as the input to the second cropper ($\ddot{I}_2 \leftarrow \dot{I}_1$). Multiple layers of croppers can handle coarse to fine detailed transformations.

\subsection{Deep Image Enhancement}\label{seq:Enhancer}

A variety of distortions, discolorations, monitor glares, deformations, etc. may exist in the input $\ddot{I}$ and/or in the cropped image $\dot{I}$. We propose to incorporate an image-to-image CNN based network to enhance the quality of $\dot{I}$ and produce the final output of the network, named $\Tilde{I} \in \mathcal{R}^{H \times W \times C}$. Image enhancement has a rich literature, and we discussed some aspects of this problem in Section~\ref{seq:rel}. However, we observe that a super-resolution network, which helps to increase the number of pixels in the image and reduce the effect of stretching pixels give us the best results. We partially adopt the architecture proposed in ~\cite{lim2017enhanced} as a base model for the enhancer sub-module of our approach. This architecture includes PixelShuffle~\cite{shi2016real} (also known as depth-to-space) to increase the resolution of images, and also has residual blocks to enable the network to produce detailed patterns from few input pixels (see Figure~\ref{fig:abstract}). Though the authors in \cite{lim2017enhanced} use $L_1$ loss to train the super resolution network, we propose a different loss in Section \ref{seq:loss} and update the parameters of the enhancer in an end-to-end fashion with the cropper.

\subsection{Loss Function} \label{seq:loss}
We formulate the \textbf{Cropper Loss} $\mathcal{L}_\mathcal{C}$ for the cropper module as the cosine distance between spatial VGG19 features (denoted by $\Gamma$ in Section~\ref{seq:DeepCropper}) of the cropped image $\dot{I}$ and the ground truth $I$:
\begin{equation} \label{eq:closs}
    \mathcal{L}_\mathcal{C} =1 - \dfrac{\Gamma(\dot{I}) \cdot \Gamma(I)}{||\Gamma(\dot{I})||^2 \cdot ||\Gamma(I)||^2},
\end{equation}
where ``$\cdot$'' operation represents the dot product, and $||.||^2$ represents the Euclidean norm. The cropper loss reduces the perceptual distance between the cropped image and the ground truth. 

For the \textbf{Enhancer Loss}, $\mathcal{L}_\mathcal{E}$, we use Mean Square Error (MSE) on top of the final output, and ground-truth spatial VGG19 features. 
\begin{equation} \label{eq:eloss}
    \mathcal{L}_\mathcal{E} = ||\Gamma(\dot{I}) - \Gamma(I)||_{2}^{2}.
\end{equation}

The final loss value we use to train the proposed network in an end-to-end fashion is $ \mathcal{L} = \mathcal{L}_{\mathcal{C}}^{n} + \mathcal{L}_{\mathcal{E}}$, where $\mathcal{L}_{\mathcal{C}}^{n}$ represents the loss function for the output of the last cropper, and $n$ is the number of stacked croppers. Note that in back-propagation, the gradients from the Enhancer affect all the layers of the model including the enhancer and croppers, and updates the $\theta_{\mathcal{C}}$ and  $\theta_{\mathcal{C}}$ (see Equations~\ref{eq:c},~\ref{eq:e}), while the gradients of croppers affect only the cropper weights,  $\theta_{\mathcal{C}}$.



\section{Experimental Setup}
\label{sec:experimental}
\begin{table*}[!ht] 
\centering
\begin{tabular}{| c || c | c || c | c | c ||c||c|c|}
\hline
 Trained on & \multicolumn{5}{c||}{DCE-1} & DCE-S & \multicolumn{2}{c|}{DCE-S + Fine-Tuned on DCE1}\\
 \hline
 Tested on & \multicolumn{2}{c||}{DCE-1} & \multicolumn{6}{c|}{DCE-2} \\
\hline
 Network & $\mathcal{C}$  & $\mathcal{E}$ &  $\mathcal{C}$  & 2  $\mathcal{C}$ &  $\mathcal{C}$ +  $\mathcal{E}$  & $\mathcal{C}$ &  $\mathcal{C}$  & 2  $\mathcal{C}$\\
 \hline
 \textbf{PSNR} & 11.36 & \textbf{16.17} & 12.34 & 12.34 & \textbf{12.68}& 12.39  & 12.52 & \textbf{12.73}\\
 \textbf{SSIM} & 0.4363 & \textbf{0.4840}  & 0.3372 & \textbf{0.3448} & 0.3213 & 0.3300 & 0.3355 & \textbf{0.3537}\\
 \textbf{MSE} & 0.0754 & \textbf{0.0284} & 0.0624 & 0.0609 & \textbf{0.0598}& 0.0621  & 0.0606 & \textbf{0.0569} \\
\hline
\end{tabular}
\caption{In this table, we show the quantitative results for experiments with different settings.  $\mathcal{C}$, 2$\mathcal{C}$, and  $\mathcal{C}$ +  $\mathcal{E}$ denote only one block of cropper, 2 stacked blocks of croppers, and the cropper + enhancer (full model) respectively. Note that we show the experiments in which the model is trained on DCE-1 and DCE-S, also the experiments in which we use DCE-1 or DCE-S as the validation set.}
\label{tab:realExp}
\end{table*}
\begin{figure*}[ht]
    \centering
    \includegraphics[width=  0.9\linewidth]{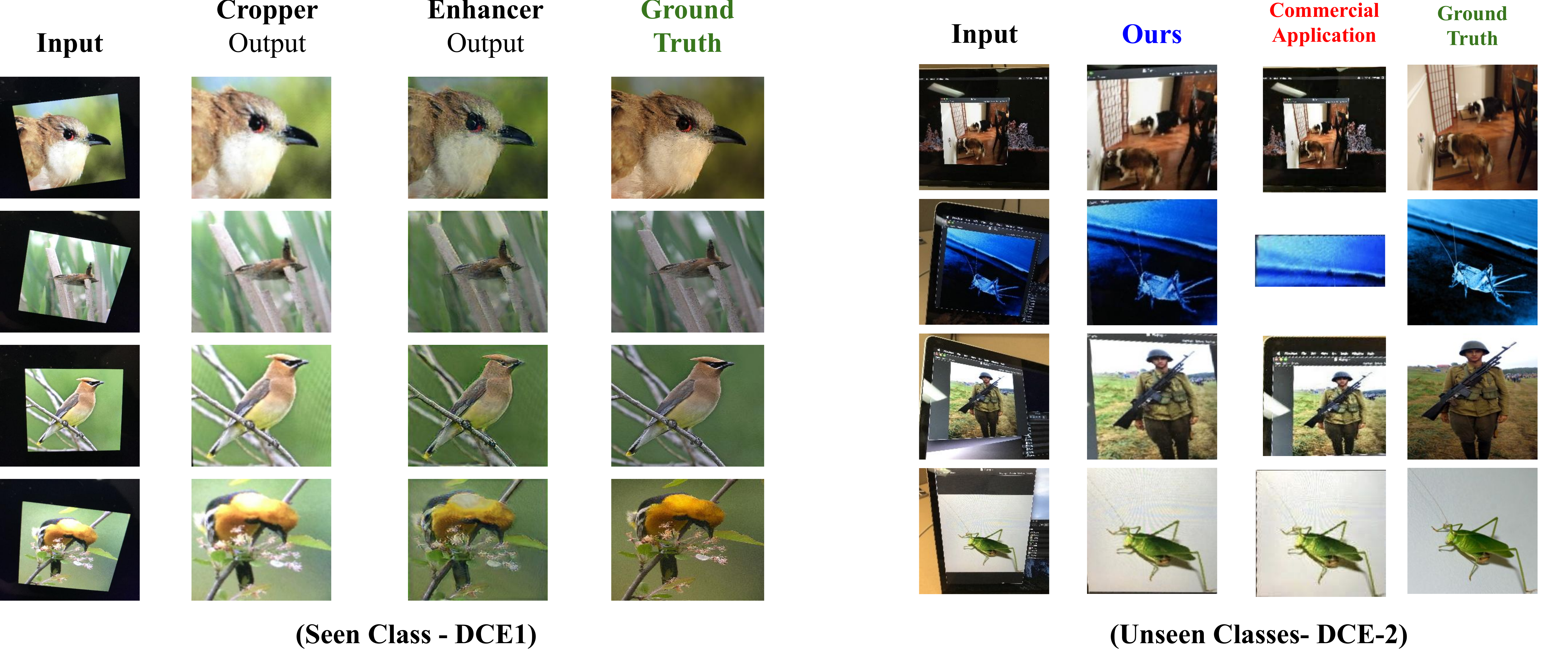}
    \caption{Qualitative results of the proposed network. In the left panel we provide qualitative examples, trained and test on DCE-1 (train and test have no common image). The Right panel shows qualitative results on DCE-2 (trained on DCE-1). Third column of the right panel shows the results out of one of the best available commercial apps. These results show that the proposed deep cropper and enhancer can successfully detect an image within a photo, and transform it to recover the original image.}
    \label{fig:DCE1}
\end{figure*}
Here, we explain our experimental setup. We describe different types of datasets we use, performance metrics, and details about the way we conduct our experiments.



To conduct this research, we collected a dataset using a smartphone camera and a monitor.   To collect the data, we randomly chose images from the Caltech-UCSD Birds 200 dataset\cite{WelinderEtal2010} as our target images. We displayed the images on a monitor and took photos at various angles and distances, while still making sure the images are large enough in the photos for details to remain distinct. This dataset is  split into two parts, which we refer to as DCE-1 and DCE-2. We make sure that the photos in DCE-2 would be more challenging than DCE-1 by putting the camera taking the pictures farther from the monitor and with more challenging backgrounds on the monitor, along with more challenging camera angles towards the monitor. Also, for the DCE-2 dataset, as shown in the right panel of Figure~\ref{fig:DCE1}, we include some images from categories other than birds (from ImageNet~\cite{deng2009imagenet}). We split the DCE-1 into train and test subsets. We only use DCE-1 training subset to train the model, while both DCE-1 test subset and DCE-2 for testing (see Table~\ref{tab:realExp}). We collected more than 100 photos for each of DCE-1 and DCE-2. We resize the input, output, and ground-truth to $224 \times 224$ for all experiments. Experiments on DCE-2 show the robustness of the model on more challenging situations than what it was trained on.





Real datasets are not easy to collect. Therefore, similar to~\cite{emrvqasongMM18,mazaheri2018visual}, we also created  a synthetic dataset. 
We took 1,000 random background images from the Places dataset~\cite{zhou2017places} and 1000 random foreground images from the Caltech-UCSD Birds 200 dataset to draw our input from. Both the foreground  and the background images were resized to 224 x 224. To generate our synthetic dataset, we start by choosing a random foreground image and applying a guided random projective transformation to it. To ensure the photos look realistic, we keep the scaling, rotation, translation, and perspective shift within certain bounds. This also allows us to make sure that the foreground image is fully in the photo. Once we have our transformed image, we embed it on a random background image. We call our synthetic dataset \textbf{DCE-S}. The foreground images are scaled down between $0.5$ and $0.8$ of the total photo size. 


\subsection{Results} \label{seq:Experiments}
We use three standard metrics, \textbf{PSNR}, \textbf{SSIM}, and \textbf{MSE} (in [0,1] scale)~\cite{ece2011image}, to measure the performance in all our experiments. In Table~\ref{tab:realExp}, we show the performance of our method in multiple scenarios. We include experiments in which we train the model on DCE-1 and DCE-S. Also, as an ablation study, for some experiments we only measure the cropper performance. We see that the enhancer can improve the results. Also, we show in the last column, that the double cropper is consistently outperforming the single cropper. This demonstrates the deep network's ability to model complicated transformations. In Figure~\ref{fig:DCE1}, we provide qualitative results of our network. We show the results on both DCE-1 and DCE-2 datasets, and also provide the output of a commercial app\footnote{https://apps.apple.com/us/app/microsoft-office-lens-pdf-scan/id975925059} in the second to last column of the right panel.N

\section{Conclusion}
\label{sec:conclusion}
In this paper, we study automatic image cropping and enhancement. We propose a deep neural network to solve this problem and discuss different aspects of designing such a network. To conduct the experiments we collected a real photos dataset, and also we proposed to create a synthetic dataset. This work introduces a proper baseline for future research on this topic.\\
\noindent \textbf{Acknowledgments:} Aaron Ott contributed to this work while he was an NSF REU student at UCF thanks to the support of NSF CNS-1757858. This work had been also supported in part by the National Science Foundation under grant number IIS-1741431.

\bibliographystyle{IEEEbib}
\bibliography{strings}

\begin{thebibliography}{10}

\bibitem{simonyan2014very}
Karen Simonyan and Andrew Zisserman,
\newblock ``Very deep convolutional networks for large-scale image
  recognition,''
\newblock {\em arXiv preprint arXiv:1409.1556}, 2014.

\bibitem{jaderberg2015spatial}
Max Jaderberg, Karen Simonyan, Andrew Zisserman, et~al.,
\newblock ``Spatial transformer networks,''
\newblock in {\em Advances in neural information processing systems}, 2015.

\bibitem{lim2017enhanced}
Bee Lim, Sanghyun Son, Heewon Kim, Seungjun Nah, and Kyoung Mu~Lee,
\newblock ``Enhanced deep residual networks for single image
  super-resolution,''
\newblock in {\em Proceedings of the IEEE conference on computer vision and
  pattern recognition workshops}, 2017.

\bibitem{shi2016real}
Wenzhe Shi, Jose Caballero, Ferenc Husz{\'a}r, Johannes Totz, Andrew~P Aitken,
  Rob Bishop, Daniel Rueckert, and Zehan Wang,
\newblock ``Real-time single image and video super-resolution using an
  efficient sub-pixel convolutional neural network,''
\newblock in {\em CVPR}, 2016.

\bibitem{yoo2019coloring}
Seungjoo Yoo, Hyojin Bahng, Sunghyo Chung, Junsoo Lee, Jaehyuk Chang, and
  Jaegul Choo,
\newblock ``Coloring with limited data: Few-shot colorization via memory
  augmented networks,''
\newblock in {\em CVPR}, 2019.

\bibitem{messaoud2018structural}
Safa Messaoud, David Forsyth, and Alexander~G Schwing,
\newblock ``Structural consistency and controllability for diverse
  colorization,''
\newblock in {\em ECCV}, 2018.

\bibitem{zhang2017real}
Richard Zhang, Jun-Yan Zhu, Phillip Isola, Xinyang Geng, Angela~S Lin, Tianhe
  Yu, and Alexei~A Efros,
\newblock ``Real-time user-guided image colorization with learned deep
  priors,''
\newblock {\em arXiv preprint arXiv:1705.02999}, 2017.

\bibitem{yeh2017semantic}
Raymond~A Yeh, Chen Chen, Teck Yian~Lim, Alexander~G Schwing, Mark
  Hasegawa-Johnson, and Minh~N Do,
\newblock ``Semantic image inpainting with deep generative models,''
\newblock in {\em CVPR}, 2017.

\bibitem{yang2017high}
Chao Yang, Xin Lu, Zhe Lin, Eli Shechtman, Oliver Wang, and Hao Li,
\newblock ``High-resolution image inpainting using multi-scale neural patch
  synthesis,''
\newblock in {\em CVPR}, 2017.

\bibitem{demir2018patch}
Ugur Demir and Gozde Unal,
\newblock ``Patch-based image inpainting with generative adversarial
  networks,''
\newblock {\em arXiv preprint arXiv:1803.07422}, 2018.

\bibitem{elad2006image}
Michael Elad and Michal Aharon,
\newblock ``Image denoising via sparse and redundant representations over
  learned dictionaries,''
\newblock {\em IEEE Transactions on Image processing}, 2006.

\bibitem{Yang_2019_CVPR}
Yang Yang, Wenye Ma, Yin Zheng, Jian-Feng Cai, and Weiyu Xu,
\newblock ``Fast single image reflection suppression via convex optimization,''
\newblock in {\em CVPR}, 2019.

\bibitem{Punnappurath_2019_CVPR}
Abhijith Punnappurath and Michael~S. Brown,
\newblock ``Reflection removal using a dual-pixel sensor,''
\newblock in {\em CVPR}, 2019.

\bibitem{Hu_2019_CVPR}
Xuecai Hu, Haoyuan Mu, Xiangyu Zhang, Zilei Wang, Tieniu Tan, and Jian Sun,
\newblock ``Meta-sr: A magnification-arbitrary network for super-resolution,''
\newblock in {\em CVPR}, 2019.

\bibitem{pathak2016context}
Deepak Pathak, Philipp Krahenbuhl, Jeff Donahue, Trevor Darrell, and Alexei~A
  Efros,
\newblock ``Context encoders: Feature learning by inpainting,''
\newblock in {\em CVPR}, 2016.

\bibitem{singh2018self}
Suriya Singh, Anil Batra, Guan Pang, Lorenzo Torresani, Saikat Basu, Manohar
  Paluri, and CV~Jawahar,
\newblock ``Self-supervised feature learning for semantic segmentation of
  overhead imagery.,''
\newblock in {\em BMVC}, 2018.

\bibitem{friston1995spatial}
Karl~J Friston, John Ashburner, Christopher~D Frith, J-B Poline, John~D
  Heather, and Richard~SJ Frackowiak,
\newblock ``Spatial registration and normalization of images,''
\newblock {\em Human brain mapping}, vol. 3, no. 3, pp. 165--189, 1995.

\bibitem{ashburner1997spatial}
John Ashburner and Karl~J Friston,
\newblock ``Spatial transformation of images,''
\newblock {\em Human brain function}, pp. 43--58, 1997.

\bibitem{davatzikos1997spatial}
Christos Davatzikos,
\newblock ``Spatial transformation and registration of brain images using
  elastically deformable models,''
\newblock {\em Computer Vision and Image Understanding}, 1997.

\bibitem{tian2003comprehensive}
Gui~Yun Tian, Duke Gledhill, and David Taylor,
\newblock ``Comprehensive interest points based imaging mosaic,''
\newblock {\em Pattern Recognition Letters}, 2003.

\bibitem{dai2017deformable}
Jifeng Dai, Haozhi Qi, Yuwen Xiong, Yi~Li, Guodong Zhang, Han Hu, and Yichen
  Wei,
\newblock ``Deformable convolutional networks,''
\newblock in {\em ICCV}, 2017.

\bibitem{He_2017_ICCV}
Kaiming He, Georgia Gkioxari, Piotr Dollar, and Ross Girshick,
\newblock ``Mask r-cnn,''
\newblock in {\em The IEEE International Conference on Computer Vision (ICCV)},
  Oct 2017.

\bibitem{glorot2010understanding}
Xavier Glorot and Yoshua Bengio,
\newblock ``Understanding the difficulty of training deep feedforward neural
  networks,''
\newblock in {\em Proceedings of the thirteenth international conference on
  artificial intelligence and statistics}, 2010.

\bibitem{WelinderEtal2010}
P.~Welinder, S.~Branson, T.~Mita, C.~Wah, F.~Schroff, S.~Belongie, and
  P.~Perona,
\newblock ``{Caltech-UCSD Birds 200},''
\newblock Tech. {R}ep. CNS-TR-2010-001, California Institute of Technology,
  2010.

\bibitem{deng2009imagenet}
Jia Deng, Wei Dong, Richard Socher, Li-Jia Li, Kai Li, and Li~Fei-Fei,
\newblock ``Imagenet: A large-scale hierarchical image database,''
\newblock in {\em CVPR}, 2009.

\bibitem{emrvqasongMM18}
Xiaomeng Song, Yucheng Shi, Xin Chen, and Yahong Han,
\newblock ``Explore multi-step reasoning in video question answering,''
\newblock in {\em Proceedings ACM-MM}, 2018.

\bibitem{mazaheri2018visual}
Amir Mazaheri and Mubarak Shah,
\newblock ``Visual text correction,''
\newblock in {\em Proceedings of the European Conference on Computer Vision
  (ECCV)}, 2018, pp. 155--171.

\bibitem{zhou2017places}
Bolei Zhou, Agata Lapedriza, Aditya Khosla, Aude Oliva, and Antonio Torralba,
\newblock ``Places: A 10 million image database for scene recognition,''
\newblock {\em IEEE Transactions on Pattern Analysis and Machine Intelligence},
  2017.

\bibitem{ece2011image}
C~Ece,
\newblock ``Image quality assessment techniques pn spatial domain,''
\newblock {\em IJCST}, vol. 2, no. 3, 2011.

\end{thebibliography}

\end{document}